\pdfoutput=1

\documentclass[11pt]{article}

\usepackage{acl}

\usepackage{times}
\usepackage{latexsym}

\usepackage{authblk}
\usepackage{multirow}

\usepackage[T1]{fontenc}

\usepackage[utf8]{inputenc}

\usepackage{microtype}

\usepackage{inconsolata}

\usepackage{amsfonts}
\usepackage{amsmath}
\usepackage{enumitem}
\usepackage{graphicx}
\usepackage{kantlipsum}
\usepackage{wrapfig}
\usepackage{comment}

\newcommand{\be}{\begin{equation}}
	\newcommand{\ee}{\end{equation}}
\newcommand{\bes}{\begin{equation*}}
	\newcommand{\ees}{\end{equation*}}

\usepackage{array}
\newcommand{\head}[2]{\multicolumn{1}{>{\centering\arraybackslash}p{#1}}{\textbf{#2}}}

\usepackage{ragged2e}

\title{Prompt Perturbation Consistency Learning for Robust Language Models}

\author[1\thanks{This work was done while interning at Amazon.}]{\textbf{Yao Qiang}}
\author[2]{\textbf{Subhrangshu Nandi}}
\author[2]{\textbf{Ninareh Mehrabi}}
\author[2]{\textbf{Greg Ver Steeg}}
\author[2]{\\\textbf{Anoop Kumar}}
\author[2,3]{\textbf{Anna Rumshisky}}
\author[2]{\textbf{Aram Galstyan}}

\affil[1]{Wayne State University, Detroit, USA} 
\affil[2]{Amazon}
\affil[3]{University of Massachusetts Lowell}

\affil[1]{\texttt {yao@wayne.edu}}
\affil[2]{\texttt {\{subhrn,mninareh,gssteeg,anooamzn,arrumshi,argalsty\}@amazon.com}}

\begin{document}
\maketitle

\begin{abstract}
Large language models (LLMs) have demonstrated impressive  performance on a number of natural language processing tasks, such as question answering and text summarization. However, their performance on sequence labeling tasks, such as intent classification  and slot filling (IC-SF), which is a central component in personal assistant systems, lags significantly behind discriminative models. Furthermore, there is a lack of substantive research on robustness of LLMs to various perturbations in the input prompts. The contributions of this paper are three-fold. First, we show that fine-tuning sufficiently large LLMs can produce IC-SF performance comparable to discriminative models. Next, we systematically analyze the  performance deterioration of those fine-tuned models due to three distinct yet relevant types of input perturbations - oronyms, synonyms, and paraphrasing. Finally, we propose an efficient mitigation approach, {\em prompt perturbation consistency learning} (PPCL), which works by regularizing the divergence between losses from clean and perturbed samples. Our experiments show that PPCL can recover on an average 59\% and 69\% of the performance drop for IC and SF tasks, respectively. Furthermore, PPCL beats data augmentation approach while using ten times fewer augmented data samples.
\end{abstract}
\section{Introduction}

\noindent
Voice controlled smart personal assistants like Amazon Echo and Google Home have flourished in recent years, enabling goal-oriented conversations and aiding tasks like setting reminders, checking weather, controlling smart devices, and online shopping. 
A core capability of those systems is to perform accurate  and robust intent classification (IC) and slot filling (SF)~\cite{tur2011spoken,qin2021survey}. The IC task involves identifying the speaker's desired intent from a given utterance, while the SF task involves recognizing the key arguments of the intent. For instance, given a user query ``wake me up at five am this week.", the intent is `set alarm', while the SF component should identify the specific details, such as `five am' as time and `this week' as date for the alarm setting. 

Pre-trained LLMs hold promise of greatly improving personal assistant systems, owing to their impressive conversational and reasoning capabilities. In addition to generating fluent conversations, LLMs have shown SOTA performance on a variety of natural language processing (NLP) tasks  such as text classification, question answering, text summarization~\cite{chowdhery2022palm,qin2023chatgpt}. Furthermore, some LLMs have shown promising ability to generate structured outputs such as code synthesis~\cite{nijkamp2023codegen2} and API calls~\cite{patil2023gorilla}. However, the performance of LLMs on other structured prediction tasks such as slot filling lags significantly behind. 

Another important issue is that LLMs can be highly sensitive to prompt variations~\cite{webson-pavlick-2022-prompt,min-etal-2022-rethinking,ye2022the}. For instance, varying  the order of few-shot examples, introducing minor typos or different expressions with the same semantic meaning can lead to qualitatively different results~\cite{jin2020bert,li2020bert,huang2021robustness,zhuo2023robustness}. In conversational systems, such perturbations might be caused by  automatic speech recognition (ASR) errors, linguistic differences, and user-specific expressions. Thus, adopting LLMs for voice-based personal assistants requires a good understanding of their robustness to above types of perturbations, and effective mitigation to have robust LLM-based IC-SF models.



\begin{figure*}[]
  \centering 
  \includegraphics[width=16cm]{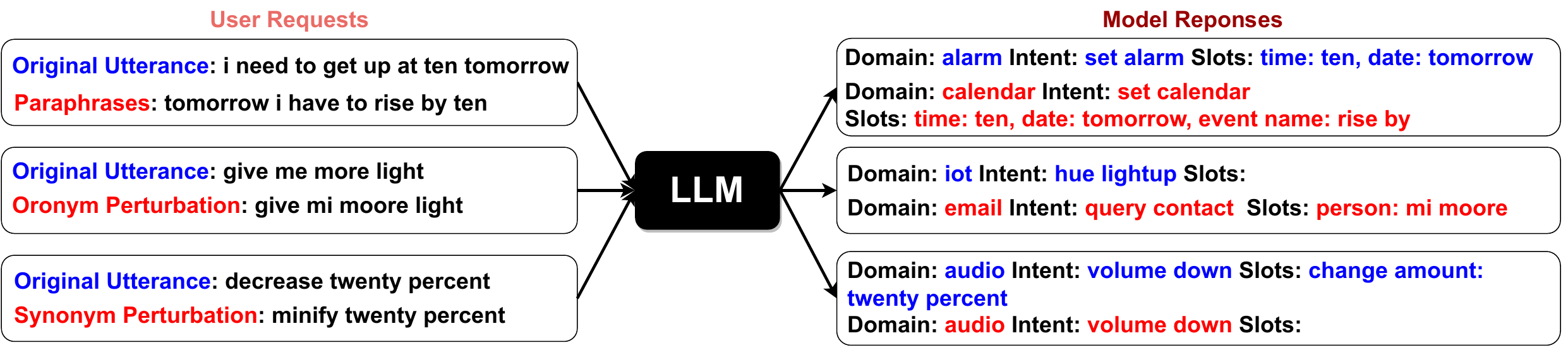}
  \caption{Illustration examples. LLMs are expected to generate structured hypotheses, i.e., domain, intent, and slots, in their responses to given user requests. Model prediction (shown in red) changes for minor perturbance. }
  \label{fig:illustration} 
\end{figure*}

In this paper we mainly consider the following questions: (1) How can we close the performance gap between LLMs and SOTA discriminative models on IC-SF tasks? (2) How does the performance of LLMs change due to minor changes in the original utterances? (3) Can we improve the robustness of LLMs in the cases of realistic perturbations?

To address the first question, we explore supervised  fine-tuning (SFT) for the IC-SF task, where the base LLM is asked to generate a target output based on an input query. We conduct extensive experiments on three publicly available NLU benchmark datasets (ATIS, SNIPS, MASSIVE) and show that by combining prompt selection and SFT on moderately sized datasets, LLMs can learn to generate structured IC-SF hypotheses with accuracy that is on par with SOTA discriminative method.  

Next, we analyze the robustness of the fine-tuned models to three different types of input perturbations that are relevant in the context of voice assistant systems -- oronyms, synonyms, and paraphrasing. 
We find that all three types of perturbations negatively impact the model performance, resulting in a significant performance drop on IC-SF tasks. 



Finally,  we propose a novel framework that we call \textit{prompt perturbation consistency learning}, or PPCL, to improve the robustness of LLMs against perturbations. Our framework (1) generates perturbed counterparts given the original utterance by either replacing a small subset of tokens or paraphrasing the utterance while constraining the semantic similarity, (2) fine-tunes LLMs with an additional consistency regularization term in the objective which explicitly encourages the model to generate consistent predictions for the original utterance and its perturbed counterpart. We conduct extensive experiments and demonstrate that PPCL can recover on an average 59\% and 69\% of the dropped performance for IC and SF tasks against perturbations, respectively. Furthermore, our results indicate that PPCL outperforms simple data augmentation approach while using only 10\% of augmented dataset.

\section{Related Work}
\vspace{-0.1cm}
\paragraph{Intent Classification and Slot Filling} 
Various techniques have been explored for intent classification\cite{Sarikaya2011,chen2012,ravuri2015recurrent}, with recent work focusing on transformer-based models and transfer learning with pre-trained language models~\cite{qin2021survey}. Slot filling, on the other hand, is typically approached using sequence labeling models, such as conditional random fields (CRFs), bidirectional LSTMs, and transformer-based architectures~\cite{weld2022survey,chen2019bert,goo2018slot,he2023can}. For a recent survey of joint IC-SF methods, see ~\cite{ic-sf-survey2022}

\vspace{-0.1cm}
\paragraph{Data Augmentation}
In NLP tasks, data augmentation methods have been explored to generate new instances by manipulating a few words in the original text~\cite{feng2021survey,chen2023empirical}. Some common techniques include word replacement, random deletion, and word position swap~\cite{wei2019eda}. Additionally, data augmentation in NLP can involve creating entirely artificial examples using back-translation~\cite{sennrich2015improving} or generative models like variational auto-encoders~\cite{malandrakis2019controlled,yoo2019data}. 
Data augmentation has also become popular for NER tasks and has been shown to be effective strategy for boosting model performance~\cite{dai2020analysis,meng2021distantly,zhou2021melm}.

\vspace{-0.1cm}
\paragraph{Consistency Training}
Consistency training methods aim to improve the robustness of models by enforcing the stability of their predictions under small perturbations, such as random noise, adversarial noise, or data augmentation techniques, applied to input examples or hidden states. 
Several attempts have been made to implement consistency training in NER tasks, utilizing both token-level and sequence-level approaches. Token-level consistency involves regularizing the model to remain unaffected by Gaussian noise~\cite{lowell2020unsupervised} or word replacement, operating at the same granularity as NER~\cite{dai2020analysis,liu2022low}. However, using such simplistic noise or augmentation methods may violate the assumption that the noised tokens should retain the same labels as the original tokens. 
Alternatively, a sequence-level consistency method employs high-quality augmentation, like back-translation, to enhance consistency across the entire sentence~\cite{xie2020unsupervised}. Nonetheless, this approach overlooks the precise location of entities due to word alignment issues, leading to a sub-optimal design. More recently, ConNER has been proposed to foster consistent predictions between a span of tokens in the original sentence and their corresponding projection in a translated sentence~\cite{zhou2022conner}. Unfortunately, ConNER's applicability is confined to cross-lingual NER tasks. Consistency training for fine-tuning LLMs on IC-SF tasks has not been thoroughly explored yet.

\noindent
\section{Method}

\subsection{Problem Formulation}
\noindent
Our main objective is to utilize LLMs for the purpose of generating structured hypotheses. As illustrated in Figure \ref{fig:illustration}, LLMs are expected to generate correct, coherent, and structured responses, including domain, intent, and slot labels, based on user utterances. To fill the performance gap between LLMs and SOTA discriminative models, we apply instruction fine-tuning~\cite{touvron2023llama}.

We decompose our task into five steps: (1) Prompts Construction: we design several prompt structures, outlined in Appendix Table \ref{tab:prompt}, to be employed during our instruction fine-tuning process. These prompts utilize the input utterances $X$ and the target outputs $Y$, which encompass various labels such as $Y_\mathrm{domain}$, $Y_\mathrm{intent}$, and $Y_\mathrm{slots}$; (2) Instruction Fine-tuning: during instruction fine-tuning, we utilize both the input ($X$) and output ($Y$) within the prompt structure, denoted as Prompt($X,Y$). This approach assists LLMs in learning the task of predicting structured hypotheses, specifically focusing on tasks like IC-SF within our investigation; (3) Response Generation: subsequent to instruction fine-tuning, we employ prompts with only input data, referred to as Prompt($X$), to elicit responses from the LLMs. These responses manifest as a generated text sequence, denoted as $W = \{w_1, \cdots, w_n\}$; (4) Obtaining Structured Hypotheses:  the generated text sequence $W$ is then transformed into structured hypotheses, culminating in the final outcomes denoted as $\{\hat{Y}_\mathrm{domain}, \hat{Y}_\mathrm{intent}, \hat{Y}_\mathrm{slots}\}$; (5) Performance Evaluation: we evaluate the performance by comparing the ground truth labels $\{Y_\mathrm{domain}, Y_\mathrm{intent}, Y_\mathrm{slots}\}$ with the outputs from the LLMs $\{\hat{Y}_\mathrm{domain}, \hat{Y}_\mathrm{intent}, \hat{Y}_\mathrm{slots}\}$. Various metrics are employed for this evaluation, e.g., accuracy and F1-score for IC and SF, respectively.

LLMs exhibit vulnerability to perturbations \cite{zhuo2023robustness,zhu2023promptbench}, leading to the generation of incorrect responses, as demonstrated in Figure \ref{fig:illustration}. Introducing small perturbations to the inputs $X$ or expressing them differently while preserving the same meaning would result in distinct inputs denoted as $X^\prime$. Nevertheless, given that $X^\prime$ maintains identical structured hypotheses and target labels $Y$, our expectation is that LLMs should be able to generate correct responses. In other words, LLMs are expected to be robust against these perturbations and generate consistent responses. 


\subsection{Prompts Construction}
\noindent
The standard prompts employed during instruction fine-tuning process with LLMs typically involve presenting both the input context and its corresponding target output in a paired structure \cite{liu2023pre}. The LLMs are then trained to generate the target output based on the input context. The primary objective here is to fine-tune the models' parameters aiming to minimize prediction errors and improve their ability to generate accurate and contextually appropriate responses. 

We construct several prompt formats for IC-SF tasks as detailed in Appendix Table \ref{tab:prompt}. The simple prompt format involves presenting the utterance and target outputs consecutively. Next, we design a structured prompt format that for predicting structured hypotheses. As shown in Appendix Table \ref{tab:prompt}, this format associates the intent with its corresponding domain and aligns the slot labels with the arguments of the request.

Furthermore, in the context of the sequence labeling task, i.e., SF, it is expected that LLMs generate slot labels for each individual token within the given utterance. Effectively associating tokens with their respective slot labels is crucial to enhance the models' performance during instruction fine-tuning. Therefore, we construct three different SF prompt formats with the intention of improving model proficiency in the SF task. The tag-only format represents the simplest approach, but it is more challenging since the model is required to implicitly track token indices as well \cite{raman2022transforming}. To simplify, we introduce sentinel-based formats. These sentinel markers enable us to avoid redundant inclusion of the original tokens in the target output. Instead, the sentinel tokens are employed to facilitate the learning of associations between tokens and their corresponding slot labels.

Our constructed prompt formats offer several advantages: (1) The structured format efficiently arranges the input and output labels within a coherent structure, facilitating the generation of structured hypotheses; (2) The sentinel-based formats eliminate the need for redundant input repetition, simplifying the decoding process and preventing hallucinations; (3) These formats enable a more straightforward method for token tracking (including indices) and establishing connections between tokens and their corresponding slot labels.

\subsection{Perturbations}
\noindent
A robust model aims to convert all utterances with or without meaning-preserving perturbations into correct hypotheses. To evaluate model robustness in IC-SF tasks, we employ different types of perturbations: oronyms, synonyms, and paraphrases, covering both word-level and sentence-level perturbations aligned with real-world application scenarios. We show some examples of these perturbations in Appendix Table \ref{tab:perturbations} and present more details of the generation process in Section \ref{sec:perturbed}.    

Oronym perturbation involves making changes to a text by replacing words or phrases with 
those 
that are phonetically similar but carry 
a different meaning.
Oronym perturbation is widely used for data augmentation in NLP tasks, especially for tasks that require robustness to speech recognition errors (ASR) or homophonic ambiguity \cite{cai2023dialogvcs}. While the altered semantics of oronym-perturbed expressions may differ from the initial utterances, our expectation is that LLMs should exhibit robustness to these changes and produce responses aligned with user intent.

Synonym perturbation replaces certain words or phrases with their 
synonyms while preserving the overall meaning of the text. It is commonly employed in NLP as data augmentation to enhance data diversity by generating new variations of a given sentence while retaining semantic coherence \cite{alfonso2021nature}. 
Synonym perturbation tests robustness of LLMs in generating consistent hypotheses when presented with semantically similar utterances.

Paraphrasing perturbation entails rephrasing a given text to create variations while preserving its original meaning. 
This is highly consistent with our daily communications that present the same meaning in different ways. Hence, irrespective of the chosen words or structures, LLMs should consistently produce accurate hypotheses.

\subsection{Data Augmentation}
\noindent
Data augmentation is widely used in fine-tuning LLMs to improve their 
generalization capabilities. There are two major benefits of 
data augmentation: (1) It expands the dataset, which proves 
beneficial for overcoming limited training data in diverse real-world scenarios; (2) It diversifies the fine-tuning dataset, equipping the model to better handle linguistic variations and consequently enhancing its performance in downstream tasks.

We apply a range of data augmentation techniques, each designed to generate diverse data through specific perturbations. 
To elaborate, we utilize word replacement techniques involving oronyms and synonyms as forms of data augmentation. This approach improves LLM's ability to adapt to previously unseen data and comprehend language variations, addressing the challenges associated with speech recognition and linguistic ambiguity. 
We also paraphrase the training data, providing LLMs with more examples to learn different ways of expressing the same content.

However, even though data augmentation is advantageous, it is essential not to introduce noise or potentially misleading content. We establish specific constraints during the generation process and implement post-processing filters to reinforce the preservation of the original utterances' integrity. 

\subsection{Prompt Perturbation Consistency Learning (PPCL)}
\noindent
Despite the fact that data augmentation has been demonstrated to be efficient to improve model robustness and generalizability \cite{chen2021hiddencut}, it overlooks the similar semantic meaning shared between the original and augmented data. To address this, we propose perturbation consistency learning framework to further utilize these augmented data, particularly the perturbed counterparts of the original utterances in our study. The key idea is to integrate a term into the training objective that explicitly encourages the generation of similar predictions (and consequently, comparable responses) for both the original utterance and its perturbed counterpart. Through the incorporation of this additional constraint, our goal is to strengthen the model's ability to maintain consistency between the original and perturbed utterances, resulting in improved robustness and more reliable performance across real-world applications.

Our objective is to align the model's responses when presented with two semantically equivalent utterances. To achieve this, we add an extra component into the training objective: the Jensen-Shannon (JS) divergence of output probabilities between a clean utterance and its perturbed counterpart. This term is integrated with the standard cross-entropy loss utilized in the auto-regression phase of the fine-tuning process.

\begin{figure}[t]
  \centering 
  \includegraphics[width=\columnwidth]{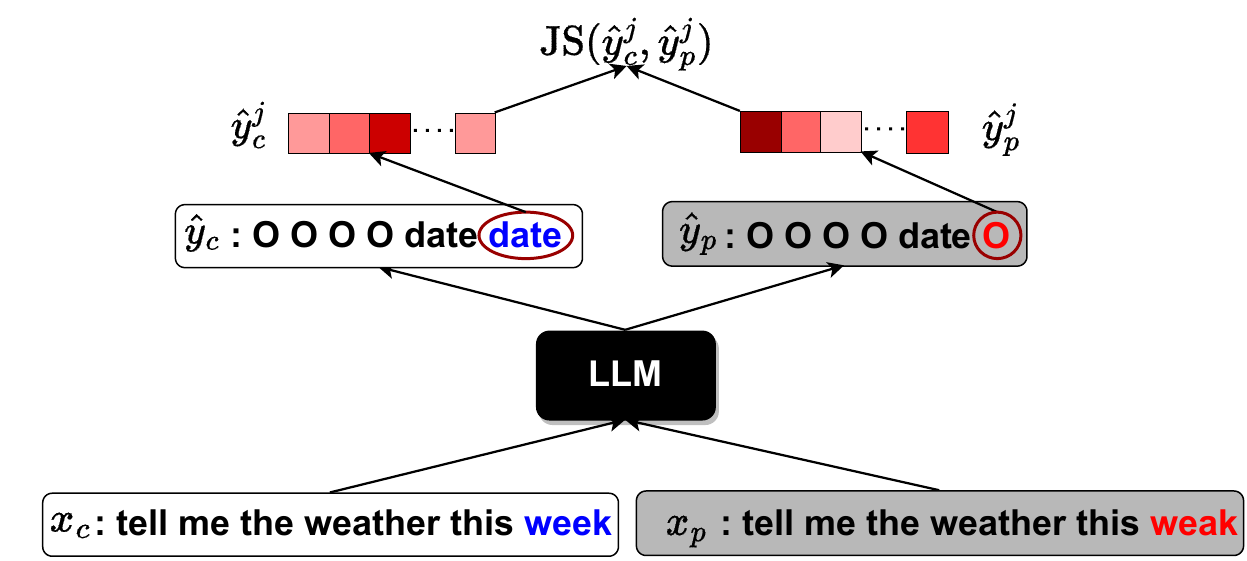} %
  \caption{Perturbation consistency learning architecture. $x_c$ and $x_p$ denote the clean and perturbed utterances, respectively. $\hat{y}_c$ and $\hat{y}_p$ here denote the slot labels generated by LLM. $\hat{y}_c^j$ and $\hat{y}_p^j$ represent the output probability distributions of current interest tokens, i.e., `date' and `O'. JS here denotes Jensen–Shannon divergence.} 
  \label{fig:pcl} 
\end{figure}

Figure \ref{fig:pcl} shows the architecture of PPCL.  During the fine-tuning process, we simultaneously input the clean utterance denoted as $x_c$ and its perturbed counterpart labeled as $x_p$ to the LLMs. In response to these inputs, the LLMs generate corresponding outputs $p_c^j$ and $p_p^j$, respectively, the probability distributions over vocabulary of the $j$-th output token for $x_c$ and $x_p$, where $p_c^j,p_p^j \in \mathbb{R}^{|\mathcal{V}|}$ and $\mathcal{V}$ denotes the vocabulary size. Subsequently, we apply Softmax to $p_c^j$ and $p_p^j$ and get their respective probability distributions $\hat{y}_c^j$ and $\hat{y}_p^j$, formally: $\hat{y}_c^j = \mathrm{Softmax}(p_c^j)$ and $\hat{y}_p^j = \mathrm{Softmax}(p_p^j)$.
%
We then apply JS divergence to quantify the similarity between $\hat{y}_c^j$ and $\hat{y}_p^j$. JS is a symmetric variation of Kullback–Leibler divergence (KL), defined as: 
\begin{equation}
    \label{eq:JSD}
    \mathrm{JS}(\hat{y}_c^j||\hat{y}_p^j) = \frac{1}{2}(\mathrm{KL}(\hat{y}_c^j||\hat{y}_m^j) + \mathrm{KL}(\hat{y}_p^j||\hat{y}_m^j)), 
\end{equation}
where $\hat{y}_m^j = \frac{1}{2}(\hat{y}_c^j + \hat{y}_p^j)$. JS smooths out the asymmetry of KL and offers a more balanced perspective on similarity. 
We obtain the JS of the two probability distributions of $j$-th output, denoted as: $\mathrm{JS}(\hat{y}_c^j\ ||\ \hat{y}_p^j)$. We use the average JS across all output probability distributions associated with $x_c$ and $x_p$ as our final perturbation consistency learning loss, formally: 
\begin{equation}
    \label{eq:L_JSD_v1}
    \mathcal{L}_\mathrm{JS} = \frac{1}{L} \sum_{j=1}^L \mathrm{JS}(\hat{y}_c^j\ ||\ \hat{y}_p^j),
\end{equation}
where $L$ denotes the response length.

Utilizing Eq. \ref{eq:L_JSD_v1} with oronym and synonym perturbations is straightforward, as these perturbations merely substitute tokens or phrases with their respective oronyms and synonyms while maintaining the utterance length. However, paraphrasing perturbations lead to varying lengths between the clean utterance and its modified counterpart. Instead of computing the JS for each token-pair in the output, we employ the averaged probability distribution to calculate the perturbation consistency learning loss for paraphrasing perturbations, formally: 

\begin{equation}
   \label{eq:L_JSD_v2}
   \mathcal{L}_\mathrm{JS} = \mathrm{JS}(\overline{\hat{y}_c}\ ||\ \overline{\hat{y}_p}),
\end{equation}

\subsection{Training Objective}
\noindent
Our training objective integrates the supervised cross-entropy losses for both clean and perturbed utterances (i.e., $\mathcal{L}_C$ and $\mathcal{L}_P$) with the perturbation consistency learning loss $\mathcal{L}_\mathrm{JS}$, formally:
\begin{equation}
   \mathcal{L}_C = \mathrm{CE}(\hat{y}_c, y), 
\end{equation}
\begin{equation}
   \mathcal{L}_P = \mathrm{CE}(\hat{y}_p, y),
\end{equation}
\begin{equation}
   \label{eq:objective}
   \mathcal{L} = \lambda_1 \mathcal{L}_C + \lambda_2 \mathcal{L}_P + \lambda_3 \mathcal{L}_\mathrm{JS},
\end{equation}
where $\lambda_1$, $\lambda_2$, and $\lambda_3$ are weight coefficients.

In order to optimize the above objective, it is essential to have both the clean utterance and its corresponding perturbed counterpart. We generate these paired perturbed utterances using our proposed perturbation generation methods. Furthermore, to ensure the presence of semantically comparable pairs, we implement specific post-processing filtering procedures. These filters serve to verify that the generated perturbed utterances genuinely maintain semantic equivalence with their clean counterparts.
\vspace{-0.15cm}

\vspace{-0.15cm}
\section{Experiments}
\subsection{Experimental Settings}
\noindent
\paragraph{Datasets}
We evaluate model performance on three NLU benchmark datasets, i.e., ATIS \cite{price1990evaluation}, SNIPS \cite{coucke2018snips}, MASSIVE \cite{fitzgerald2022massive}. More details of these datasets and their statistics are shown in the Appendix. 

\paragraph{Prompt Formats}
We show our proposed prompt formats with an illustrated example for IC-SF tasks in Table \ref{tab:prompt}. 

\begin{table*}[h]
\centering
\small
\caption{Illustration of prompt and SF formats for IC-SF tasks}
\label{tab:prompt}
\begin{tabular}{lllllllll}
\cline{1-2}
\multicolumn{2}{l}{{\color[HTML]{333333} \begin{tabular}[c]{@{}l@{}}Utterance (\textbf{\textcolor{red}{u}}): wake me up at five am this week Domain (\textbf{\textcolor{blue}{d}}): alarm Intent (\textbf{\textcolor{brown}{i}}): alarm\_set\\ Slots (\textbf{\textcolor{orange}{s}}): {[}Other Other Other Other time time date date{]}  Arguments (\textbf{\textcolor{olive}{a}}): {[}time : five am, date : this week{]}\end{tabular}}}                                     &  &  &  &  &  &  &  \\ \cline{1-2}
\multicolumn{1}{l}{\textbf{Prompt Format}}                      & \textbf{Samples}                                                                                                                                                                                                                                                   &  &  &  &  &  &  &  \\ 
\multicolumn{1}{l}{Simple Prompt}                               & Utterance: \textbf{\textcolor{red}{u}} Domain: \textbf{\textcolor{blue}{d}} Intent: \textbf{\textcolor{brown}{i}} Slots: \textbf{\textcolor{orange}{s}} Arguments: \textbf{\textcolor{olive}{a}}                                                                                                                                                                                                             &  &  &  &  &  &  &  \\ 
\multicolumn{1}{l}{Structured Prompt}                           & Utterance: \textbf{\textcolor{red}{u}} Intent in Domain: \textbf{\textcolor{brown}{i}} in \textbf{\textcolor{blue}{d}} Slots with Arguments: \textbf{\textcolor{orange}{s}} with \textbf{\textcolor{olive}{a}}                                                                                                                                                                                               &  &  &  &  &  &  &  \\ \cline{1-2}
\multicolumn{1}{l}{\textbf{SF Format}}                         & \textbf{Sample Inputs \& Slots}                                                                                                                                                                                                                                    &  &  &  &  &  &  &  \\ 
\multicolumn{1}{l}{}                                            & Input: wake me up at five am this week                                                                                                                                                                                                                             &  &  &  &  &  &  &  \\ 
\multicolumn{1}{l}{\multirow{-2}{*}{Tag Only}}                  & Slots: Other Other Other Other time time date date                                                                                                                                                                                                                 &  &  &  &  &  &  &  \\ 
\multicolumn{1}{l}{}                                            & Input: \textless{}0\textgreater{}wake \textless{}1\textgreater{}me \textless{}2\textgreater{}up \textless{}3\textgreater{}at \textless{}4\textgreater{}five \textless{}5\textgreater{}am \textless{}6\textgreater{}this \textless{}7\textgreater{}week             &  &  &  &  &  &  &  \\ 
\multicolumn{1}{l}{\multirow{-2}{*}{Sentinel + Tag}}            & Slots: \textless{}0\textgreater{}Other \textless{}1\textgreater{}Other \textless{}2\textgreater{}Other \textless{}3\textgreater{}Other \textless{}4\textgreater{}time \textless{}5\textgreater{}time \textless{}6\textgreater{}date \textless{}7\textgreater{}date &  &  &  &  &  &  &  \\ 
\multicolumn{1}{l}{}                                            & Input: \textless{}0\textgreater{}wake \textless{}1\textgreater{}me \textless{}2\textgreater{}up \textless{}3\textgreater{}at \textless{}4\textgreater{}five \textless{}5\textgreater{}am \textless{}6\textgreater{}this \textless{}7\textgreater{}week             &  &  &  &  &  &  &  \\ 
\multicolumn{1}{l}{\multirow{-2}{*}{Extractive Sentinel + Tag}} & Slots: \textless{}4\textgreater{}time \textless{}5\textgreater{}time \textless{}6\textgreater{}date \textless{}7\textgreater{}date                                                                                                                                 &  &  &  &  &  &  &  \\ \cline{1-2}
\end{tabular}
\end{table*}

\paragraph{Baselines} 
We compare the performance of PPCL with the following baselines: supervised fine-tuning with discriminative models like JointBERT and JointBERT+CRF, zero-shot and few-shot learning with GPT variants, instruction fine-tuning with LLaMA. For additional information about these baselines and their specific experimental setups, please refer to the Appendix.

\subsection{Evaluation Metrics}
\noindent
For the IC task, we use prediction accuracy on a held-out test set, and for the SF task, we use the F1-score as the evaluation metrics. Instead of using absolute differences in performance between models trained with clean and perturbed data, we use a relative measurement. We introduce Performance Drop Rate (PDR), which quantifies the relative performance decline following a perturbation, formally: 
\begin{equation}
    \mathrm{PDR}(\mathcal{D},\mathcal{D}^\prime,f_\theta) = 1 - \frac {\sum_{(x,y)\in{\mathcal{D^\prime}}}\mathcal{M}[f_\theta(x),y]}{\sum_{(x,y)\in{\mathcal{D}}}
\mathcal{M}[f_\theta(x),y]}.
\end{equation}
$\mathcal{M}$ here is the indicator function and $f_\theta$ denotes the models. $\mathcal{D}$ and $\mathcal{D}^\prime$ indicates the clean and perturbed test sets, respectively. We want to clarify that the clean and perturbed test sets are in a one-to-one correspondence, thus $|\mathcal{D}| == |\mathcal{D}^\prime|$. In other words, each example in the clean test set has a corresponding example in the perturbed test set. This ensures a fair and direct comparison between the model's performance on clean and perturbed samples.

\vspace{-0.25cm}
\subsection{Perturbed Evaluation Sets}
\label{sec:perturbed}
\noindent
We generate perturbed evaluation sets for each benchmark dataset. The synonym perturbation involves randomly choosing and substituting words with their synonyms based on the WordNet synonym corpus. The oronym perturbation follows a similar procedure relying on the NLTK pronouncing corpus. Specifically, we compile a list of key stop words based on the domain, intent, and slot label sets, and do not substitute them. Additionally, we have imposed a limit of three words as the maximum number that can be perturbed in an utterance to prevent significant changes in semantic meaning. We generate the paraphrases using a specific LLM from \href{https://huggingface.co/humarin/chatgpt_paraphraser_on_T5_base}{Huggingface}, which is specially pre-trained for generating high-quality paraphrases. To further ensure that clean and perturbed samples are semantically similar, we filter out perturbations with BERTScore \cite{zhang2019bertscore} with the original sample. We use a 0.85 threshold based on our empirical experimental studies.

With perturbations of samples, generating appropriate target labels is crucial for evaluation. For intent labels, we align them with those of the original utterances. For slot labels, the procedure is more complex. For perturbations that maintain the length and word order, such as oronyms and synonyms, we directly adopt the original slot labels as their corresponding counterparts. For paraphrased variations that may deviate in length and word order from the original utterance, we automatically generate new slot labels. The new slot labels are derived from the semantic annotations present in the original utterance. This strategy ensures that the perturbed versions retain their intended meaning while accommodating any structural changes arising from the paraphrasing process.


\section{Results and Discussion}

\begin{table}[t]
\centering
\small
\caption{Comparison of model performance on three datasets. The best performance of SOTA discriminative models and LLMs is highlighted in bold.}
\label{tab:baselines}
\scalebox{0.95}{
\begin{tabular}{llcc}
\hline
\textbf{Datasets} & \textbf{Model} & \textbf{Intent Acc} & \textbf{Slot F1} \\
\hline
 & JointBERT & \textbf{89.44} & 80.43 \\
 & JointBERT+CRF & 88.67 & \textbf{80.58} \\
 & GPT3.5-ZS & 60.39 & - \\
MASSIVE & GPT3.5-FS & 67.18 & 31.76 \\
 & GPT2+SFT & 84.13 & 66.72 \\
 & LLaMA-7b+SFT & 88.01 & 80.45 \\
 & LLaMA-13b+SFT & 88.87 & 80.7 \\
 & LLaMA-30b+SFT & \textbf{89.05} & \textbf{80.74} \\
\hline
 & JointBERT & \textbf{97.53} & \textbf{95.83} \\
 & JointBERT+CRF & 96.75 & 95.58 \\
ATIS & GPT3.5-ZS & 87.45 & - \\
 & GPT3.5-FS & 93.17 & 73.51 \\
 & GPT2+SFT & 97.31 & 83.92 \\
 & LLaMA-7b+SFT & \textbf{98.21} & \textbf{94.26} \\
\hline
 & JointBERT & \textbf{98.57} & \textbf{96.67} \\
 & JointBERT+CRF & 98.28 & 96.07 \\
SNIPS & GPT3.5-ZS & 95.14 & - \\
 & GPT3.5-FS & 94.42 & 49.12 \\
 & GPT2+SFT & 97.14 & 88.23 \\
 & LLaMA-7b+SFT & \textbf{98.14} & \textbf{94.51} \\
\hline
\end{tabular}}
\footnotetext[1]{CRF: Conditional Random Field; FS: Few Shot; SFT: Supervised Fine Tuned; ZS: Zero Shot}
\vspace{-0.25cm}
\end{table}


\subsection{Performance Gap between LLMs and discriminative models}
\noindent
First, we show the model performance comparison of different baselines on three datasets in Table \ref{tab:baselines}. These results demonstrate that LLMs, i.e., GPT2 and LLaMA, which have been instruction fine-tuned with our proposed sentinel-based structured format, achieve comparable intent classification performance to SOTA discriminative models like JointBERT across all three datasets. However, applying zero-shot and few-shot learning settings  the performance of LLMs is notably worse, especially for the SF tasks.

The lower performance of LLMs on the SF task could be attributed to the mismatch between the nature of the semantic labeling task and the design of text generation models. 
The latter are not inherently optimized for SF tasks, which might lead to sub-optimal results in some cases. However they can still achieve comparable results for the sequence labeling task, such as SF, after supervised fine-tuning with appropriate instructions or structured formats. This is demonstrated by LLaMA-30b achieving and average SF accuracy (89.84\%) within 1.3\% of JointBERT performance (91.03\%), and even superseding it for MASSIVE dataset.

It is important to highlight that the key advantage of using generative models over discriminative models for IC-SF tasks lies in their ability to create and understand a wider range of linguistic variations. Generative models can generate new examples, enhancing the training set with diverse phrases and structures. This leads to a more robust model that can better handle varied user inputs. In contrast, discriminative models typically rely on the existing training set, which might limit their ability to adapt to new or unexpected ways people express similar intents.

\subsection{Prompt Formats}
\noindent
We compare the model performance using different prompt formats in Table \ref{tab:promptformats}. 
The sentinel-based structured prompt format achieves the best performance, particularly for the SF tasks. This outcome aligns with our initial hypothesis that the structured format is highly effective in organizing both the input and output labels, leading to improved learning ability for the models. In addition, sentinel-based slot formatting significantly improves performance. 

\begin{table}[t]
\centering
\small
\caption{Comparison of model performance with different prompt formats: Simple and Structured prompt formats with tag-only, extractive sentinel-based with tag, and sentinel-based with tag slots formats, respectively.}
\label{tab:promptformats}
\scalebox{0.85}{
\begin{tabular}{llll}
\hline
\head{1cm}{\textbf{Datasets}} & \head{4cm}{\textbf{Prompt Formats}} & \head{0.6cm}{\textbf{Intent Acc}} & \head{0.5cm}{\textbf{Slot F1}} \\
\hline
 & Simple + Tag & \textbf{98.43} & 86.04 \\
ATIS & Simple + Extractive Sentinel & 97.76 & 93.12\\
 & Simple + Sentinel Tag & 98.21 & \textbf{94.26} \\
\hline
 & Simple + Tag & 97.85 & 89.11 \\
SNIPS & Simple + Extractive Sentinel & \textbf{98.71}  & 92.88 \\
& Simple + Sentinel Tag & 98.14 & \textbf{94.51} \\
\hline
 & Simple + Tag & 88.68 & 72.91 \\
 & Simple + Extractive Sentinel & 88.33 & 73.42 \\
 & Simple + Sentinel Tag & 87.51 & 75.36 \\
MASSIVE & Structured + Tag & \textbf{88.73} & 75.72 \\ 
 & Structured + Extractive Sentinel & 87.82 & 75.13 \\
 & Structured + Sentinel & 88.01 & \textbf{80.45} \\
\hline
\end{tabular}}
\end{table}

\subsection{Performance Drop due to Prompt Perturbations}

\begin{table*}[t]
\centering
\small
\caption{Comparison of model performance drops as a result of prompt perturbations, on MASSIVE dataset. The smaller PDR values imply higher model robustness.}
\label{tab:pdrs}
\begin{tabular}{llrrrrrr}
\hline
\textbf{Perturb} & \textbf{Model} & \textbf{Clean IC} & \textbf{Perutbed IC} & \textbf{IC-PDR} & \textbf{Clean SF} & \textbf{Perturbed SF} & \textbf{SF-PDR} \\
\hline
 & JointBERT & 90.19 & 70.77 & 21.53 & 80.50& 42.28 & 47.47 \\
 & JointBERT+CRF & 89.50 & 71.19 & 20.45 & 80.65 & 42.41 & 47.41 \\
 & GPT3.5-ZS & 61.39 & 60.69 & 1.15 & - & - & - \\
Oronyms  & GPT3.5-FS & 70.43 & 48.91 & 30.55 & 31.95 & 20.75 & 35.05 \\
 & GPT2+SFT & 85.52 & 67.71 & 20.83 & 65.14 & 27.51 & 58.40 \\
 & LLaMA-7b+SFT & 89.18 & 74.31 & 16.67 & 79.35 & 47.01 & 40.75 \\
\hline
 & JointBERT & 90.43 & 78.29 & 13.42 & 80.83 &74.77 & 7.49 \\
 & JointBERT+CRF & 89.43 & 77.61 & 13.21 &81.86 & 75.87 & 7.31 \\
 & GPT3.5-ZS & 63.04 & 58.66 & 6.95 & - & - & - \\
Synonyms  & GPT3.5-FS & 65.54 & 54.59 & 16.71 & 34.43 & 31.57 & 8.30 \\
 & GPT2+SFT & 84.99 & 70.42 & 17.14 & 67.92 & 60.62 & 10.74 \\
 & LLaMA-7b+SFT & 89.23 & 76.79 & 13.94 & 80.75 & 72.90 & 9.72 \\
\hline
 & JointBERT & 89.30 & 82.96 & 7.09 & 82.81 & 71.67 & 13.45 \\
 & JointBERT+CRF & 88.71 & 80.88 & 8.82 & 82.64 & 70.08 & 15.19 \\
 & GPT3.5-ZS &60.80 & 55.27 & 9.09 & - & - & - \\
Paraphrases  & GPT3.5-FS & 65.55 & 59.08 & 9.88 & 34.87 & 29.22 & 16.20 \\
 & GPT2+SFT & 82.60 & 76.71 & 7.13 & 63.53 & 52.33 & 17.63 \\
 & LLaMA-7b+SFT & 82.78 & 80.21 & 8.62 & 81.58 & 68.41 & 16.14 \\
\hline
\end{tabular}
\end{table*}

\begin{table*}
\centering
\small
\caption{Mitigation results of data augmentation and PPCL on MASSIVE dataset. We show results with different augmentation sizes and different loss functions. For multi-sample augmentation the training size increase by $\sim 50k$, for single sample it is similar to the original size.}
\label{tab:ablation}
\scalebox{0.94}{
\begin{tabular}{llclrrrr}
\hline
\textbf{Perturb} & \textbf{Mitigation} & \textbf{Augmentation} & \textbf{Loss} & \textbf{IC-PDR} & \textbf{Recovery} & \textbf{SF-PDR} & \textbf{Recovery} \\
\hline
 & Baseline & - & $\mathcal{L}_c$ & 16.67 & - & 40.75 & - \\
 & JS Loss & +3k & $\mathcal{L}_c+\mathcal{L}_{js}$ & 15.74 & 5\% & 32.80 & 19\%\\
Oronyms & Perturb Loss & +3k & $\mathcal{L}_c+\mathcal{L}_p$ & 8.95 & 46\% & 18.44 & 55\% \\
 & Perturb Loss & +50k & $\mathcal{L}_c+\mathcal{L}_p$ & 9.02 & 45\% & 19.73 & 51\% \\
 & PPCL (JS + Perturb Loss) & +3k & $\mathcal{L}_c+\mathcal{L}_p +\mathcal{L}_{js}$  & \textbf{8.74} & \textbf{47\%} & \textbf{15.41} & \textbf{62\%} \\
\hline
 & Baseline & - & $\mathcal{L}_c$ & 13.94 & - & 9.72 & - \\
 & JS Loss & +5k & $\mathcal{L}_c+\mathcal{L}_{js}$ & 12.11 & 13\% & 7.83 & 19\%\\
Synonyms & Perturb Loss & +5k & $\mathcal{L}_c+\mathcal{L}_p$ & 5.59 & 60\% & 5.13 & 47\% \\
 & Perturb Loss & +50k & $\mathcal{L}_c+\mathcal{L}_p$ & 4.01 & 71\% & 4.49 & 53\% \\ 
 & PPCL (JS + Perturb Loss) & +5k & $\mathcal{L}_c+\mathcal{L}_p +\mathcal{L}_{js}$ & \textbf{3.74} & \textbf{73\%} & \textbf{1.44} & \textbf{85\%} \\
\hline
 & Baseline & - & $\mathcal{L}_c$ & 8.62 & - & 16.14 & - \\
 & JS Loss & +6k & $\mathcal{L}_c+\mathcal{L}_{js}$ & 7.79 & 9\% & 15.10 & 6\% \\
Paraphrases & Perturb Loss & +6k & $\mathcal{L}_c+\mathcal{L}_p$ & 5.92 & 31\% & 8.89 & 45\% \\
 & Perturb Loss & +50k & $\mathcal{L}_c+\mathcal{L}_p$ & \textbf{3.69} & \textbf{57\%} & \textbf{4.24} & \textbf{74\%} \\
 & PPCL (JS + Perturb Loss) & +6k & $\mathcal{L}_c+\mathcal{L}_p+\mathcal{L}_{js}$ & \textbf{3.69} & \textbf{57\%} & 6.36 & 60\% \\
\hline
\end{tabular}}
\end{table*}

\noindent
Table \ref{tab:cases} illustrates examples of clean and perturbed utterances and their difference in model predictions even though the BertScores between the clean and perturbed samples are higher than 0.85. We show the relative performance drops resulting from the following three perturbations: oronyms, synonyms, and paraphrases, on MASSIVE dataset in Table \ref{tab:pdrs}. The results of ATIS and SNIPS are shown in Appendix. Results show that discriminative models, ICL approaches, and LLMs with instruction fine-tuning are vulnerable to these perturbations with large performance drops, most notably, in SF tasks with oronym perturbations. 

These findings highlight the vulnerabilities of both discriminative and generative models when exposed to perturbed data, emphasizing the need to improve model robustness for real-world applications.
Identifying and mitigating the impact of perturbations, especially in tasks involving sequence labeling like SF, are critical to improving the performance and generalizability of these models.

\begin{table*}[t]
\centering
\small
\caption{Some examples of clean and perturbed utterances, with BertScore $>$ 0.85. Red lines are a result of perturbation. Blue lines are post PPCL mitigation.}
\label{tab:cases}
\begin{tabular}{lllll}
\hline
Perturbations & Utterances                          & Pred\_Domain & Pred\_Intent        & Pred\_Slots                          \\ \hline
Clean      & create an alarm for today at ten am & alarm        & alarm\_set          & {[}today: date , ten am: time{]} \\
\textcolor{red}{Paraphrase}    & \textcolor{red}{set a reminder for today at ten am}  & \textcolor{red}{calendar}     & \textcolor{red}{calendar\_set}       & \textcolor{red}{{[}today: date , ten am: time{]}} \\ 
\textcolor{blue}{Paraphrase}    & \textcolor{blue}{set a reminder for today at ten am}  & \textcolor{blue}{alarm}        & \textcolor{blue}{alarm\_set}          & \textcolor{blue}{{[}today: date , ten am: time{]}}   \\ \hline
Clean      & give   me more lite                 & iot          & iot\_hue\_lightup   & {[}{]}  \\ 
\textcolor{red}{Oronym}        & \textcolor{red}{give mi moore lite}              & \textcolor{red}{email}        & \textcolor{red}{email\_querycontact} & \textcolor{red}{{[}mi moore: person{]}} \\ 
\textcolor{blue}{Oronym}        & \textcolor{blue}{give mi moore lite}                  & \textcolor{blue}{iot}          & \textcolor{blue}{iot\_hue\_lightup}   & \textcolor{blue}{{[}{]}}     \\ \hline
\end{tabular}
\end{table*}

\subsection{PPCL Mitigation Results}
\noindent
We share results from two mitigation approaches for improving robustness of LLMs against prompt perturbations: data augmentation and PPCL. We show results with different augmentation sizes and different combinations of loss functions on MASSIVE dataset are in Table \ref{tab:ablation}. All these are done on LLaMA-7b model. Both approaches decrease the significant performance drop. The ones where multiple perturbed samples are added for each clean sample the training data size increases by 50k or more. For example, data augmentation with one perturbed sample per clean sample, along with perturbation loss, shown as $\mathcal{L}_C + \mathcal{L}_P$ recovers performance drops up to 45\% on IC and 51\% on SF tasks, respectively for Oronym perturbation. When augmented with 5 perturbed samples per clean sample, it performs better. However, PPCL, with only 1 perturbed sample per clean, which includes perturbation loss and JS loss, outperforms multiple sample augmentation in all cases, except for SF in paraphrase perturbation. For paraphrase perturbation, PPCL recovers 60\% of SF-PDR compared to 74\% by multi-sample augmentation, but at one-tenth the augmentation size. On average, PPCL is able to recover 59\% in IC and 69\% in SF performance drops. In comparison, multi-sample augmentation is able to recover 58\% in IC and 59\% in SF. PPCL achieves the recoveries with one-tenth the augmentation size. PPCL comparisons with augmentation on ATIS and SNIPS datasets as shown in Appendix, indicating the generalizability and effectiveness of our approach across different domains and datasets.

\subsection{Ablation Studies}
\noindent
In our training objective, there are three different terms in Eq. \ref{eq:objective}, and to better understand their contributions towards improving the robustness of LLMs against perturbations, we conducted an ablation study as shown in Table \ref{tab:ablation}.  Experimental results make it clear
that the models achieve the best performance when all three loss terms ($\mathcal{L}_c,\ \mathcal{L}_p, \ \mathcal{L}_{js}$) in the training objective are utilized, indicating each term plays a significant role in enhancing the robustness of the models. PPCL outperforms multi-sample augmentation with a fraction of augmentation volume in 5 out of 6 tasks in Massive data. 

We have also carefully fine-tuned the three weights in the PPCL loss (Eq. \ref{eq:objective}) for each dataset respectively to identify the best-performing model. To improve model performance, we believe that these weights should be carefully fine-tuned and selected under different settings and datasets.

\subsection{Failure and Saved Examples}
We provide two case studies in Table \ref{tab:cases} to illustrate some failure due to the perturbations and the recoveries after applying PPCL. In these two examples, we observe that oronym substitution and paraphrasing lead the model to generate incorrect responses. These incorrect responses (red lines) are characterized as failure cases, as they do not accurately capture the user's intents or the relevant information in the utterances. However, after re-training the model with PPCL, we see improvement. The model is now able to generate the correct responses, which are demonstrated in blue lines.

\vspace{-0.15cm}
\section{Conclusion}
\vspace{-0.15cm}
\noindent
We study, evaluate, and improve the robustness of LLMs in generating structured hypotheses, such as IC-SF tasks. We first propose a sentinel-based structured prompt format for instruction fine-tuning LLMs resulting in comparable performance to SOTA discriminative models. Next, we evaluate robustness of LLMs under various prompt perturbations, i.e., synonyms, oronyms, and paraphrases. Our results indicate that LLMs are vulnerable to these perturbations, with an average performance drop rate of 13.07\% in IC accuracy and 22.20\% in SF F1-score. We then propose two mitigation strategies, i.e., perturbation consistency learning and data augmentation, aiming to improve model robustness. These methods can recover up to 59\% performance drop in IC task and 69\% in SF task, making the resulting LLMs more robust to prompt perturbations. Finally, our findings show that PPCL surpasses the basic data augmentation method, achieving superior performance with just 10\% of the augmented datasets, thereby exhibiting enhanced scalability.

\section*{Limitations}

PPCL was developed based on observations on publicly available small datasets like Massive, ATIS, SNIPS. The improvement in performance might not be as pronounced in real world datasets whose distributions and noise structure might not mimic the public datasets. Improvement in robustness by implementing PPCL was evaluated on IC-SF tasks. We expect PPCL to work in other tasks as well, but we have not demonstrated it. We plan to do so in future work. 
\section*{Ethics Statement}
The authors foresee no ethical concerns with the research presented in this work. They also completed an internal legal review process which verified that we are using publicly available models and datasets consistent with their intended use.

\section*{Acknowledgements}
The authors would like to thank the reviewers and area chairs for their suggestions and comments.



\clearpage
\bibliography{anthology,custom}

\appendix

\label{sec:appendix}

\clearpage
\section{Appendix}

\subsection{Datasets}
\label{appendix:datasets}
We show the data statistics of the three datasets in Table \ref{tab:dataset} and present more details here. 

\noindent
\textbf{ATIS}: ATIS dataset has been widely used to develop and evaluate natural language understanding systems, including intent detection, slot-filling, and dialogue act classification. The dataset consists of a collection of human-computer dialogues, where users interact with a simulated airline information system to obtain various travel-related information, such as flight schedules, ticket availability, and airport information. These dialogues were collected from real users interacting with the ATIS system.

\noindent
\textbf{SNIPS}: SNIPS dataset is designed to support the development and evaluation of voice-controlled systems for home automation tasks. It consists of a large collection of spoken language interactions, where users interact with a voice assistant to perform various tasks commonly found in a home setting, such as setting alarms, playing music, checking the weather, and controlling smart devices.

\noindent
\textbf{MASSIVE}: MASSIVE dataset is an open source multilingual NLU dataset from Amazon Alexa NLU systems consisting of 1 million labeled utterances spanning 51 language. For our experiments, we only use the en-US domain utterances.

\begin{table*}[h]
\centering
\small
\caption{Dataset statistics}
\label{tab:dataset}
\begin{tabular}{lccccc}
\hline
\textbf{Datasets} & \textbf{Train} & \textbf{Dev} & \textbf{Test} & \textbf{Intent Labels} & \textbf{Slot Labels} \\
\hline
ATIS & 4478 & 500 & 893 & 18 & 127 \\
SNIPS & 13084 & 700 & 700 & 7 & 72 \\
MASSIVE & 11514 & 2033 & 2974 & 60 & 56 \\
\hline
\end{tabular}
\end{table*}

\subsection{Baselines}

\noindent
\textbf{JointBERT and JointBERT+CRF}: JointBERT was propose in \cite{chen2019bert} as a joint IC-SF model based on BERT. JointBERT+CRF investigates the efficacy of adding Conditional Random Field (CRF) for modeling slot label dependencies on top of the joint BERT model. We use English uncased BERT-Base model which has 12 layers, 768 hidden states, and 12 heads. For fine-tuning, all hyper-parameters are tuned on the development set. The maximum length is 50. The batch size is 32. Adam is used for optimization with an initial learning rate of 5e-5. The dropout probability is 0.1. The maximum number of epochs is set as 10.

\noindent
\textbf{Zero/Few-shot Learning}: In our experiments, we utilize the OpenAI API and GPT3.5 for conducting zero-shot and few-shot learning tasks. We use 10 examples in the few-shot learning. Different prompts are designed to evaluate the model's ability to generalize and perform tasks it hasn't been explicitly trained on, showcasing its capacity for zero-shot and few-shot learning scenarios.

\noindent
\textbf{LLMs}: We evaluate several popular LLMs, including GPT-2 and LLaMA. GPT-2 is a large-scale unsupervised language model designed to generate human-like text based on the context given to it. We use the smallerst version of GPT-2 with 124M parameters. The LLaMA model is a collection of foundation language models ranging from 7B to 65B parameters proposed by Meta. We use the 7b, 13b, and 30b versions during our experiments.

\noindent
\textbf{Supervised Fine-tuning}: We first apply supervised fine-tuning with LLMs for IC-SF tasks. The maximum length is set as 256. The batch size is 32. Adam is also use for optimization with an initial learning rate of 3e-4 with 100 steps warm-up. We fine-tune the model 5 ecpochs.

\noindent
\textbf{Perturbation Consistency Learning}: We further fine-tune the models for another 2 epochs with out perturbation consistency learning objective. We use Adam as optimizer with an initial learning rate of 3e-4.

\subsection{Perturbation Examples}
We show several examples of different types of perturbations in Table \ref{tab:perturbations}.

\begin{table*}[h]
\centering
\small
\caption{Examples of different types of perturbations}
\label{tab:perturbations}
\begin{tabular}{c|c}
\hline
\textbf{Original Utterances}    & \textbf{Oronyms Perturbations}           \\ 
review all alarms                & review aul alarms                        \\ 
when is the event going to start & wynn is the event going to start         \\ \hline
\textbf{Original Utterances}    & \textbf{Synonyms Perturbations}          \\ \hline
email to new contact             & email to novel contact                   \\ 
pink is all we need              & pink is all we ask                       \\ \hline
\textbf{Original Utterances}    & \textbf{Paraphrasing Perturbations}      \\ 
tell me the weather this week    & whats the weather forecast for this week \\ 
how old is mariah carey          & what is the age of mariah carey          \\ \hline
\end{tabular}
\end{table*}

\subsection{More Results}
We show some other results in the following tables. Table \ref{tab:pdrs_atis} and Table \ref{tab:pdrs_snips} show the comparison of model performance drops against different types of perturbations on ATIS and SNIPS datasets, respectively. Table \ref{tab:ablation_atis} and Table \ref{tab:ablation_snips} show the ablation studies on the different terms in training objective $\mathcal{L}$ (Eq. \ref{eq:objective}) on ATIS and SNIPS datasets, respectively. 

\begin{table*}[h]
\centering
\small
\caption{Comparison of model performance drops against perturbations on ATIS dataset.}
\label{tab:pdrs_atis}
\begin{tabular}{llrrrrrr}
\hline
\textbf{Perturb} & \textbf{Model} & \textbf{Clean IC} & \textbf{Perutbed IC} & \textbf{IC-PDR} & \textbf{Clean SF} & \textbf{Perturbed SF} & \textbf{SF-PDR} \\
\hline
 & JointBERT & 97.87 & 96.11 & 1.79 & 96.47 & 78.37 & 18.76 \\
 & JointBERT+CRF & 97.17 & 95.75 & 1.46 & 96.00 & 76.09 & 20.74 \\
 & GPT3.5-ZS & 87.80 & 86.21 & 1.81 & - & - & - \\
Oronyms  & GPT3.5-FS & 91.54 & 90.28 & 1.37 & 77.89 & 51.42 & 33.98 \\
 & GPT2+SFT & 98.58 & 96.28 & 2.33 & 59.75 & 43.49 & 27.21 \\
 & LLaMA-7b+SFT & 99.11 & 97.17 & 1.95 & 94.24 & 76.68 & 18.63 \\
\hline
 & JointBERT & 97.91 & 91.96 & 6.07 & 93.18 & 92.64 & 3.68 \\
 & JointBERT+CRF & 97.32 & 89.28 & 8.26 & 96.28 & 92.46 & 3.96 \\
 & GPT3.5-ZS & 82.44 & 76.48 & 7.22 & - & - & - \\
Synonyms  & GPT3.5-FS & 89.58 & 88.09 & 1.66 & 77.50 & 73.08 & 5.70 \\
 & GPT2+SFT & 97.32 & 92.56 & 4.89 & 60.17 & 53.00 & 11.91 \\
 & LLaMA-7b+SFT & 98.21 & 91.36 & 6.97 & 94.73 & 89.33 & 5.70 \\
\hline
 & JointBERT & 97.60 & 91.00 & 6.76 & 95.86 & 82.64 & 13.79 \\
 & JointBERT+CRF &98.81 & 90.20 & 8.71 & 95.61 & 82.43 & 13.78 \\
 & GPT3.5-ZS & 88.15 & 82.33 & 6.71 & - & - & - \\
Paraphrases  & GPT3.5-FS & 90.20 & 87.12 & 3.41 & 77.50 & 70.01 & 9.66 \\
 & GPT2+SFT & 92.12 & 90.19 & 2.09 & 92.96 & 44.76 & 51.85 \\
 & LLaMA-7b+SFT &98.17 & 90.42 & 7.89 &93.72 & 80.63 & 13.97 \\
\hline
\end{tabular}
\end{table*}

\begin{table*}[h]
\centering
\small
\caption{Comparison of model performance drops against perturbations on SNIPS dataset.}
\label{tab:pdrs_snips}
\begin{tabular}{llrrrrrr}
\hline
\textbf{Perturb} & \textbf{Model} & \textbf{Clean IC} & \textbf{Perutbed IC} & \textbf{IC-PDR} & \textbf{Clean SF} & \textbf{Perturbed SF} & \textbf{SF-PDR} \\
\hline
 & JointBERT & 98.61 & 96.06 & 2.58 & 97.05 & 79.14 & 18.45 \\
 & JointBERT+CRF & 98.14 & 94.67 & 3.53 & 95.87 & 78.63 & 17.98 \\
 & GPT3.5-ZS & 95.60 & 94.44 & 1.21 & - & - & - \\
Oronyms  & GPT3.5-FS & 93.98 & 90.74 & 3.44 & 50.30 & 41.48 & 17.53 \\
 & GPT2+SFT & 97.86 & 95.26 & 2.65 & 90.66 & 65.24 & 28.04 \\
 & LLaMA-7b+SFT & 98.14 & 96.75 & 1.42 & 94.42 & 75.84 & 19.67 \\
\hline
 & JointBERT & 99.05 & 95.58 & 3.50 & 96.00 & 87.04 & 9.33 \\
 & JointBERT+CRF & 99.05 & 95.58 & 3.50 & 94.87 & 86.68 & 8.63 \\
 & GPT3.5-ZS & 95.89 & 84.85 & 11.51 & - & - & - \\
Synonyms  & GPT3.5-FS & 94.32 & 80.44 & 14.71 & 48.05 & 43.28 & 9.92 \\
 & GPT2+SFT & 98.71 & 90.06 & 8.76 & 90.85 & 75.41 & 16.99 \\
 & LLaMA-7b+SFT & 99.05 & 94.32 & 4.77 & 94.45 & 83.25 & 11.85 \\
\hline
 & JointBERT & 98.53 & 93.09 & 5.52 & 96.67 & 58.69 & 39.39 \\
 & JointBERT+CRF & 98.23 & 91.77 & 6.57 & 96.06 & 58.88 & 38.70 \\
 & GPT3.5-ZS & 95.74 & 83.84 & 12.42 & - & - & - \\
Paraphrases  & GPT3.5-FS & 93.97 & 80.76 & 14.05 & 49.49 & 33.01 & 33.29 \\
 & GPT2+SFT & 97.60 & 90.09 & 7.69 &90.96 & 49.44 & 45.64 \\
 & LLaMA-7b+SFT &98.23 & 90.01 & 8.36 & 94.41 & 55.64 & 41.06 \\
\hline
\end{tabular}
\end{table*}

\begin{table*}[h]
\centering
\small
\caption{Ablation studies on the different terms in training objective $\mathcal{L}$ of SNIPS dataset.}
\label{tab:ablation_snips}
\begin{tabular}{llrrrr}
\hline
\textbf{Perturb} & \textbf{Losses} & \textbf{IC-PDR} & \textbf{Recovery} & \textbf{SF-PDR} & \textbf{Recovery} \\
\hline
 & $\mathcal{L}_C$ & 1.42 & - & 19.67 & - \\
Oronyms & $\mathcal{L}_C+\mathcal{L}_P$ & 0.23 & 84\% & 2.62 &86\%\\
 & $\mathcal{L}_C+\mathcal{L}_P+\mathcal{L}_{JS}$ & $\boldsymbol{0.0}$ & $\boldsymbol{100\%}$ & $\boldsymbol{1.58}$ & $\boldsymbol{92\%}$ \\
\hline
 & $\mathcal{L}_C$ & 4.77 & - & 11.85 & - \\
Synonyms & $\mathcal{L}_C+\mathcal{L}_P$ & 1.70 & 64\% & 3.89 & 67\%\\
 & $\mathcal{L}_C+\mathcal{L}_P+\mathcal{L}_{JS}$ & $\boldsymbol{+0.31}$ & $\boldsymbol{118\%}$ & $\boldsymbol{1.31}$ & $\boldsymbol{89\%}$ \\
\hline
 & $\mathcal{L}_C$ & 8.36 & - & 41.06 & - \\
Paraphrases & $\mathcal{L}_C+\mathcal{L}_P$ & 5.52 & 34\% & 28.97 & 29\% \\
 & $\mathcal{L}_C+\mathcal{L}_P+\mathcal{L}_{JS}$ & $\boldsymbol{4.63}$ & $\boldsymbol{44\%}$ & $\boldsymbol{28.45}$ & $\boldsymbol{30\%}$ \\
\hline
\end{tabular}
\end{table*}

\begin{table*}[h]
\centering
\small
\caption{Ablation studies on the different terms in training objective $\mathcal{L}$ of ATIS dataset.}
\label{tab:ablation_atis}
\begin{tabular}{llrrrr}
\hline
\textbf{Perturb} & \textbf{Losses} & \textbf{IC-PDR} & \textbf{Recovery} & \textbf{SF-PDR} & \textbf{Recovery} \\
\hline
 & $\mathcal{L}_C$ & 1.95 & - & 18.63 & - \\
Oronyms & $\mathcal{L}_C+\mathcal{L}_P$ & 0.18 & 83\% & +0.33 &101\%\\
 & $\mathcal{L}_C+\mathcal{L}_P+\mathcal{L}_{JS}$ & $\boldsymbol{+0.01}$ & $\boldsymbol{100\%}$ & $\boldsymbol{+0.71}$ & $\boldsymbol{104\%}$ \\
\hline
 & $\mathcal{L}_C$ & 6.97 & - & 5.70 & - \\
Synonyms & $\mathcal{L}_C+\mathcal{L}_P$ & 3.55 & 49\% & 2.32 & 59\%\\
 & $\mathcal{L}_C+\mathcal{L}_P+\mathcal{L}_{JS}$ & $\boldsymbol{2.11}$ & $\boldsymbol{69\%}$ & $\boldsymbol{0.33}$ & $\boldsymbol{94\%}$ \\
\hline
 & $\mathcal{L}_C$ & 7.89 & - & 13.97 & - \\
Paraphrases & $\mathcal{L}_C+\mathcal{L}_P$ & 6.51 & 17\% & 8.95 & 36\% \\
 & $\mathcal{L}_C+\mathcal{L}_P+\mathcal{L}_{JS}$ & $\boldsymbol{4.83}$ & $\boldsymbol{39\%}$ & $\boldsymbol{3.19}$ & $\boldsymbol{77\%}$ \\
\hline
\end{tabular}
\end{table*}

\end{document}